\documentclass{article}

\usepackage{arxiv}
\usepackage[utf8]{inputenc} 
\usepackage[T1]{fontenc}    
\usepackage{hyperref}       
\usepackage{url}            
\usepackage{booktabs}       
\usepackage{amsfonts}       
\usepackage{nicefrac}       
\usepackage{microtype}      
\usepackage{lipsum}		
\usepackage{upgreek}
\usepackage{graphicx}     
\usepackage{subcaption}
\usepackage{amsmath}
\newtheorem{remark}{Remark}

\title{A hypothesis-driven method based on machine learning for neuroimaging data analysis}

\author{
  J.M. Gorriz\thanks{Part of the data used in the preparation of this article were obtained
from the Alzheimer's Disease Neuroimaging Initiative (ADNI)
database (http://www.loni.ucla.edu/ADNI). As such, the investigators
within the ADNI contributed to the design and implementation of ADNI
and/or provided data but did not participate in analysis or writing of
this manuscript. ADNI investigators include (complete listing available at
http://www.loni.ucla.edu/ADNI/Collaboration/ADNI Manuscript Citations.pdf)} \\
  DASCI Institute\\
  University of Granada\\
  Granada, Spain \\
   \texttt{gorriz@ugr.es,jg825@cam.ac.uk} \\
   \And
 R.  Martin Clemente\\
  Dpt.  Signal Theory and Communications\\
  University of Seville\\
  Seville, Spain\\
  \texttt{ruben@us.es} \\
     \And
 C.G.  Puntonet\\
  Dpt.  Computer Architecture and Tech.\\
  University of Granada\\
 Granada, Spain\\
  \texttt{carlos@atc.ugr.es} \\
     \And
 A.  Ortiz\\
  Dpt. Communication Engineering\\
  University of Malaga\\
  Malaga, Spain\\
  \texttt{aortiz@ic.uma.es} \\
       \And
 J.  Ramirez\\
  DASCI Institute\\
  University of Granada\\
  Granada, Spain\\
  \texttt{javierrp@ugr.es} \\
     \And
 John Suckling \\
  Department of Psychiatry\\
  University of Cambridge\\
  Cambridge, UK\\
  \texttt{js369@cam.ac.uk} \\
}

\begin{document}
\maketitle

\begin{abstract}
There remains an open question about the usefulness and the interpretation of Machine learning (MLE) approaches for discrimination of spatial patterns of brain images between samples or activation states. In the last few decades, these approaches have limited their operation to feature extraction and linear classification tasks for between-group inference. In this context, statistical inference is assessed by randomly permuting image labels or by the use of random effect models that consider between-subject variability. These multivariate MLE-based statistical pipelines, whilst potentially more effective for detecting activations than hypotheses-driven methods, have lost their mathematical elegance, ease of interpretation, and spatial localization of the ubiquitous General linear Model (GLM).  Recently, the estimation of the conventional GLM has been demonstrated to be connected to an univariate classification task when the design matrix is expressed as a binary indicator matrix. In this paper we explore the complete connection between the univariate GLM and MLE \emph{regressions}. To this purpose we derive a refined statistical test with the GLM based on the parameters obtained by a linear Support Vector Regression (SVR) in the \emph{inverse} problem (SVR-iGLM). Subsequently, random field theory (RFT) is employed for assessing statistical significance following a conventional GLM benchmark. Experimental results demonstrate how parameter estimations derived from each model (mainly GLM and SVR) result in different experimental design estimates that are significantly related to the predefined functional task. Moreover, using real data from a multisite initiative the proposed MLE-based inference demonstrates statistical power and the control of false positives, outperforming the regular GLM.
\end{abstract}


\section{Introduction}

Whole-brain analyses in neuroimaging, comprising a large number of independent statistical tests, have been traditionally conducted with classical statistics, either hypothesis testing or Bayesian inference, and the univariate General Linear Model (GLM) \cite{Friston95}.  These hypothesis-driven methods gained their popularity due to the ease of interpretation and function localization across experimental designs \cite{Friston02}.  However, they usually rely on assumptions that are frequently violated; e.g. homogeneity, Gaussianity, etc. \cite{Rosenblatt14} and, consequentially, inflated type I error rates have become problematic and a key contributor to the replication crisis \cite{Eklund16,Noble20}. Furthermore, technological advances are increasing spatial and temporal resolutions as well as the range of available measurements of anatomy and physiology; a true exemplar of the \emph{curse of dimensionality} \cite{Bellman03}. In this context, analyses of contemporary large image repositories retain the difficulties associated with small sample sizes. One of them is the inflated false-positives observed across experiments as a consequence of the multiple comparison problem that is partly solved by over conservative approaches, such as Bonferroni or Random Field Theory (RFT) corrections \cite{Frackowiak04}. 

One promising solution is machine learning (MLE), where high-dimensional relationships between datasets are empirically established \cite{Vapnik82}.  Estimating dependencies in regression or classification tasks using  statistical learning theory (SLT), unlike classical statistics, characterizes the actual relationships or effects with a limited dataset. Neuroimaging in particular has embraced MLE as a technology to deliver diagnostic and prognostic classification \cite{Bzdok17,Illan12,Zhang14} of neurological and psychiatric disorders. Nevertheless, the mainstay of neuroimaging studies are observational and mechanistic, seeking to identify regional between-group differences in brain structure and function.  

Data-driven analysis methods based on MLE \cite{Mouro-Miranda05,Wang09,Wang07, Smith13,Gorriz2021} have demonstrated their ability for detecting activations in fMRI data, outperforming conventional hypothesis-driven approaches,  i.e.  the standard GLM inference based on random effect models. The core idea on these agnostic (model-free) approaches is to perform an accurate feature extraction based on a fixed-complexity MLE classifier, such as a linear support vector machine (SVM), between predefined groups. They all share the same characteristic processing pipeline of a data-driven multivariate approach that enhances  detection ability. Consequently,  the Statistical Parametric Maps (SPM) derived from the GLM are replaced by a spatial discriminance map (SDM) \cite{Mouro-Miranda05} based on prevalence \cite{Rosenblatt16}, or some other specific feature extracted at the training stage; e.g. distance to the separating hyperplane \cite{Wang09}. These data-driven maps are then deployed in  conventional pipelines for statistical inference, although some  approaches depart from the p-value-based frequentist, Bayesian or permutation analyses, and introduce the concept of the probability of the \emph{worst case} in neuroimaging \cite{Gorriz2021}

Despite the popularity of MLE as a solution for a wide range of complex problems, there remains an open question about its usefulness for statistical inference.  As an example, the inclusion of covariates and nuisance variables is still an issue, beyond hybrid approaches that make use of a fitting process in the ``MLE space'' in combination with the GLM.  Efforts with MLE around this issue are increasing with continuous output variables (\cite{Cohen2011}, with remarks in \cite{Reiss15}) rather than the more typical categorical classifications.  However, the classification task is just a particular case of the regression problem with continuous labels, thus exploring general MLE methods for linear regression, such as SV regression (SVR), is currently relevant in addition to its use as a simple extension of  SVM \cite{Zhang14}.

Recently, a connection between both domains, that is, the standard GLM and the inverse GLM, has been established \cite{Gorriz2021b} using binary experimental design matrices. The aim was to formalize the relationship that has been evidenced in several neuroimaging applications in the extant literature. In this paper, we show a complete and novel connection between the classical GLM, including random effect models, and the MLE framework for the estimation of optimum regression parameters using SVR in the inverse domain (SVR-iGLM). Subsequent analyses based on frequentist inference and permutation testing are carried out to calculate the level of significance in between-group testing of SDMs with a multiple comparisons correction. 

\section{Theory}\label{sec:methods}

\subsection{The General Linear Model and its statistical framework}\label{sec:GLM}

The GLM \cite{Friston02} is defined for a single observation level, e.g. in a between-subject comparison, as:
\begin{equation}\label{eq:1}
\mathbf{y}=\mathbf{X} \boldsymbol{\uptheta} + \boldsymbol{\upepsilon}
\end{equation}
where $\mathbf{y}$ is the $N\times 1$ observation vector with units of time, signal change, volume, etc., $\boldsymbol{\upepsilon}$ is the $N\times 1$ vector of errors that is assumed to be Gaussian distributed, $\mathbf{X}$ is the $N \times M$ matrix containing the explanatory variables or constraints, and $\boldsymbol{\uptheta}$ is the $M\times 1$ vector of parameters explaining the observations $\mathbf{y}$. Note that: i) for a hierarchical observation model each level requires prior estimation at the previous levels; and ii) in terms of MLE, $\mathbf{X}$ are the multidimensional labels or regressors acting on the observations $\mathbf{y}$.  In the classic GLM, $\boldsymbol{\uptheta}$ is usually estimated by a Maximum Likelihood (ML) criterion based on the Gaussianity assumption and is given by:
\begin{equation}\label{eq:2}
\hat{\boldsymbol{\uptheta}}=(\mathbf{X}^t \mathbf{C}_{\epsilon}^{-1} \mathbf{X})^{-1}\mathbf{X}^t \mathbf{C}_{\epsilon}^{-1}\mathbf{y}
\end{equation}
where $\mathbf{C}_{\epsilon}$ is the covariance matrix of errors. Inferences on this estimate\footnote{Here, we refer to voxelwise inference since we use a threshold $u$ to classify voxels $i$ as ``active'' if $T_i\geq u$.  Clusterwise inference uses a cluster-forming threshold to define contiguous suprathreshold regions \cite{Nichols12}.} determine the components of $\boldsymbol{\uptheta}$, and the relationship between classical GLM and MLE-based prevalence inferences can be obtained using a linear compound specified by a contrast weight vector $\mathbf{c}$, and writing a T statistic as:
\begin{equation}\label{eq:3}
T=\frac{\mathbf{c}^t\hat{\boldsymbol{\uptheta}} }{\sqrt{\mathbf{c}^tCov(\hat{\boldsymbol{\uptheta}})\mathbf{c} }}
\end{equation}
where $Cov(\hat{\boldsymbol{\uptheta}})=(\mathbf{X}^t \mathbf{C}_{\epsilon}^{-1} \mathbf{X})^{-1}$. This T statistic gives us the probability of observing the ML estimation under $H_0$, and when it is small enough, e.g.  $p<0.05$, the linear compound is considered significantly different from zero. As an example, given a set of two parameters in $\boldsymbol{\uptheta}=[\theta_1, \theta_2]^t$, if we select $\mathbf{c}=[1 -1]$ we are assessing how large is the first parameter with respect to the second; i.e. the difference  $\theta_1-\theta_2$. Thus, if the T statistic suggests a small probability, the contrast is statistically significant with observations generated from different sources. 

A similar procedure could be established based on a Bayesian estimation and inference to handle complex hierarchical observational models. This framework would be based on the Expectation Maximization (EM) algorithm for parameter estimation, along with known priors and \emph{a priori} probability models, with the aim of evaluating the posterior probability (ppm). By thresholding the ppm, relationships between this and the frequentist approach can be established for both their similarities (statistical power) and differences (specificity) \cite{Friston02}.

\subsection{Machine learning and the inverse GLM}
\label{sec:GLMleast}
\begin{figure}
\centering
\includegraphics[width=0.5\textwidth]{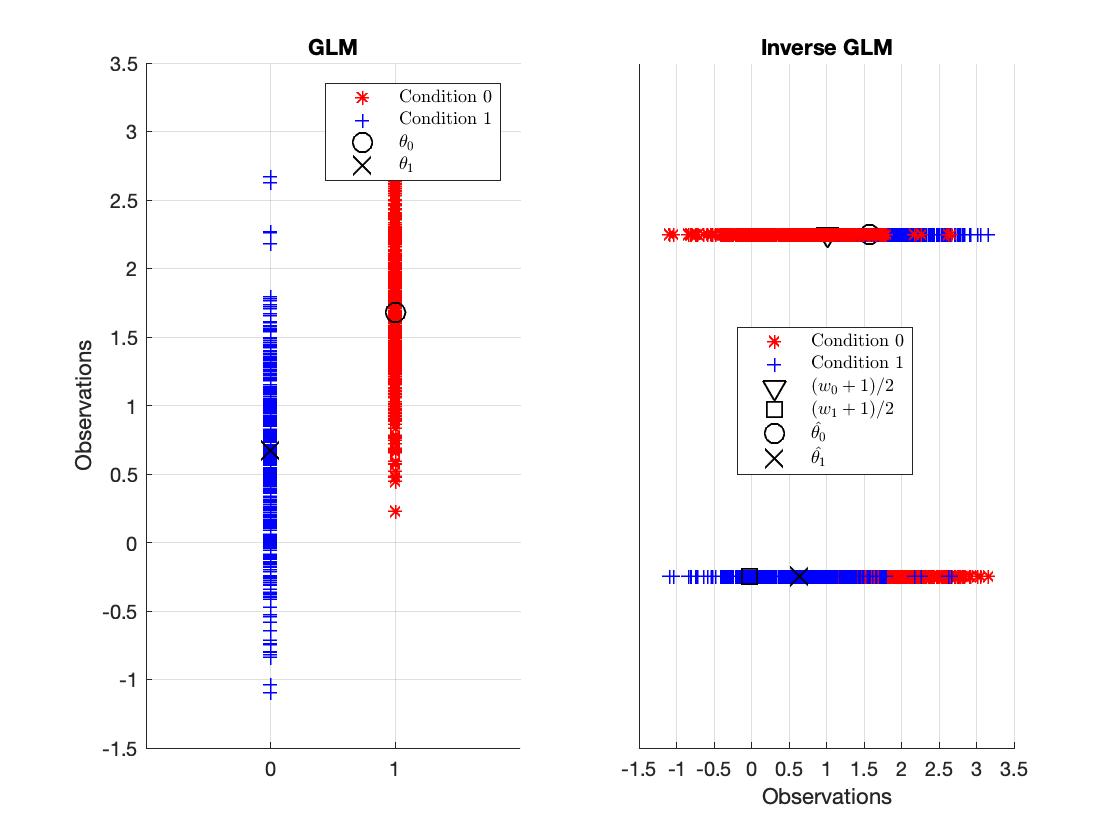}
\caption{Illustration example of the connection between GLM and iGLM. }
\label{fig:cero}
\end{figure}

The GLM can be interpreted as the inverse problem of regressing the observations onto the conditions (see figure \ref{fig:cero}). Instead of assuming the model in equation \ref{eq:1}, the inverse GLM is defined as:
\begin{equation}\label{eq:4}
\mathbf{X} =\mathbf{y}\boldsymbol{\omega}+\hat{\boldsymbol{\upepsilon}}
\end{equation}
where $\boldsymbol{\omega}$ is a set of  ($1\times M$) parameters that best explains the design matrix given the observations, and $\hat{\boldsymbol{\upepsilon}}$ is noise with unknown pdf.  We can readily see that:
\begin{equation}\label{eq:5}
(\mathbf{X}-\hat{\boldsymbol{\upepsilon}})\boldsymbol{\omega}^t (\boldsymbol{\omega}\boldsymbol{\omega}^t)^{-1}=\mathbf{y}
\end{equation}
where we assume that the inverse of the norm $(\boldsymbol{\omega}\boldsymbol{\omega}^t)^{-1}$ exists. After some manipulations we finally get:
\begin{equation}\label{eq:6}
\mathbf{y}=\mathbf{X}\tilde{\boldsymbol{\uptheta}}+\tilde{\boldsymbol{\upepsilon}}
\end{equation}
where we define:
\begin{equation}\label{eq:7}
\tilde{\boldsymbol{\uptheta}}=\boldsymbol{\omega}^t (\boldsymbol{\omega}\boldsymbol{\omega}^t)^{-1};
\quad \tilde{\boldsymbol{\upepsilon}}=-\hat{\boldsymbol{\upepsilon}}\boldsymbol{\omega}^t (\boldsymbol{\omega}\boldsymbol{\omega}^t)^{-1}
\end{equation}
Therefore, solving the multiple regression problem in equation $\ref{eq:4}$, e.g. using a MLE approach, is equivalent to estimating the parameters of the GLM. If we carefully examine equation \ref{eq:7}, each column $c$ of the design matrix ($c=1,\ldots,M$ including covariates and nuisance variables) can be described by:
\begin{equation}\label{eq:8}
\left[
\begin{array}{c}
x_{1,c} \\
\vdots \\
x_{N,c}
\end{array}
\right]= \mathbf{y}\omega_c+ \hat{\boldsymbol{\upepsilon}}_c
\end{equation}
The set of parameters $\boldsymbol{\omega}$ could be determined by a classical approach such as Parametric Empirical Bayes (PEB), assuming a Gaussian model for the noise $\hat{\boldsymbol{\upepsilon}}$. However, we prefer to use the MLE approach based on the regularized risk minimization and linear regressors.  
For the reader's convenience, further mathematical discussion is available on~\ref{Appendix1}. 

Thus, given the set of observations $\mathbf{y}$ and the experimental conditions and covariates in $\mathbf{X}$, we regress $M$ independent linear equations as:
\begin{equation}\label{eq:9}
\mathbf{X} =\mathbf{y}\boldsymbol{w}+\mathbf{B}=\hat{\mathbf{y}}\hat{\boldsymbol{w}}
\end{equation}
where $\hat{\boldsymbol{w}}=\left[
\begin{array}{c}
\boldsymbol{\omega}\\
\mathbf{B}
\end{array}
\right]$, the $N\times M$ matrix $\mathbf{B}=[\mathbf{b}^t\ldots \mathbf{b}^t]^t$ and $\mathbf{b}$ is the $1\times M$ vector of biases.  In view of equation  \ref{eq:8}, we realistically assume under this approach that the bias does not depend on the experimental condition or covariate realization.  

Once the set of parameters and biases $\{\mathbf{w},\mathbf{b}\}$ are estimated by a suitable procedure (see Appendix), such as least squares (LS) or SVR, we can calculate the observation $\mathbf{y}$ by simply inverting equation \ref{eq:9} as:
\begin{equation}\label{eq:10}
\mathbf{y}_{est}=(\mathbf{X}-\mathbf{B})\boldsymbol{w}^t(\boldsymbol{w}\boldsymbol{w}^t)^{-1}=\tilde{\mathbf{X}}\tilde{\boldsymbol{\uptheta}}
\end{equation} 
where $\tilde{\boldsymbol{\uptheta}}$ is the MLE estimated vector of parameters that combines the weight of multiple predictors to explain the observation, and $\tilde{\mathbf{X}}=(\mathbf{X}-\mathbf{B})$ is the adjusted design matrix. Instabilities that could arise from the scalar inversion in equation \ref{eq:10}, that depend on the selected MLE algorithm, are easily solved by bounding its value between $-1$ and $1$.

It is worth mentioning here that our model does not use any data reduction techniques, e.g. principal component analysis (PCA),  preserving the function localization of the  univariate GLM.  However, a similar description of the method can be given in terms of the difference of signed distances by replacing the main reference function in the GLM by $\frac{\mathbf{y}\boldsymbol{w}}{||\boldsymbol{w}||^2}$. In this way we a give an explanation of the multivariate approach proposed in \cite{Wang09} in terms of the solution in the iGLM. 

\subsection{Support Vector Regression}
In general, the core idea of SVR \cite{Scholkopf02} is to do (non-)linear regression in a feature space $\mathcal{F}$: 
\begin{equation}\label{eq:11}
f(\mathbf{y}) =\boldsymbol{w}\cdot \Phi(\mathbf{y}) +b,\quad \Phi:\mathbb{R}^n \rightarrow\mathcal{F}, \quad \boldsymbol{w}\in \mathcal{F}
\end{equation}
where $\cdot $ denotes dot product and, in the case of the linear regression, $\Phi$ is simply the identity function.  We determine $\boldsymbol{w}$ from the data by minimizing the sum of the empirical risk, e.g. $\epsilon$-insensitive loss function \cite{Burges98}, and a complexity term proportional to its norm: 
\begin{equation}\label{eq:12}
R_{reg}(f)=\frac{1}{2}||\boldsymbol{w}||^2+C\sum_{i=1}^N |x_i-f(\mathbf{y}_i)|_\epsilon
\end{equation}
This minimization can be transformed into a uniquely solvable quadratic programming problem \cite{Smola98} that provides the vector of parameters $\boldsymbol{w}$ in terms of the samples or support vectors:
\begin{equation}\label{eq:13}
\boldsymbol{w}=\sum_{i=1}^{N}(\alpha_i-\alpha_i^*)\Phi(\mathbf{y}_i)
\end{equation}
where $\alpha_i,\alpha_i^*$ are Lagrange multipliers; that is, the solutions of the quadratic programming problem.  Finally, the bias term $b$ can be computed by determining the prediction error on the margin $\delta_i=f(\mathbf{y}_i)-x_i=\epsilon \text{sign}(\alpha_i-\alpha_i^*)$ and taking the average of differences as $b=<f(\mathbf{y}_i)-\boldsymbol{w}\cdot \Phi(\mathbf{y}_i)>$.

\subsection{Model Equivalence}

After performing the regression in the iGLM domain by SVR and assuming that the explanatory matrix $\mathbf{X}$ contains two experimental conditions, i.e. indicator variables that refer to class membership $\mathbf{X}_b$ and a set of continuous covariates or nuisance variables $\mathbf{X}_c$, the observations can be estimated by:
\begin{equation}\label{eq:14}
\mathbf{y}_{est}=\mathbf{X}_b\tilde{\boldsymbol{\uptheta}}_b+\mathbf{X}_c\tilde{\boldsymbol{\uptheta}}_c-\mathbf{B}\tilde{\boldsymbol{\uptheta}}
\end{equation}
where $\tilde{\boldsymbol{\uptheta}}=[\tilde{\boldsymbol{\uptheta}}_b^t \tilde{\boldsymbol{\uptheta}}_c^t]^t$. Equation \ref{eq:10} can be used to approximate a set of parameters that best explain the observation vector with maximum and minimum influence of covariates and nuisance variables ranged in the interval $[0,1]$:
\begin{equation}\label{eq:15}
\tilde{\boldsymbol{\uptheta}}_{est}=\tilde{\boldsymbol{\uptheta}}_b+[0,1]\cdot\tilde{\boldsymbol{\uptheta}}_c-\mathbf{B}\tilde{\boldsymbol{\uptheta}}
\end{equation}
  
\section{Materials and Methods}

\subsection{Synthetic data}

We generated synthetic data with the aim of modeling different scenarios in an fMRI time-series analysis with inserted activations using a block-design (baseline and task) paradigm \cite{Wang09}. For this purpose, we simulated one dimensional observation vectors $\mathbf{y}$ (Eq \ref{eq:1}) with different contrast to noise ratios (CNR: $0.25,0.5, 0.75, 1$) and sample sizes ($N$:100 - 1000).  

\begin{figure}
\centering
\includegraphics[width=0.5\textwidth]{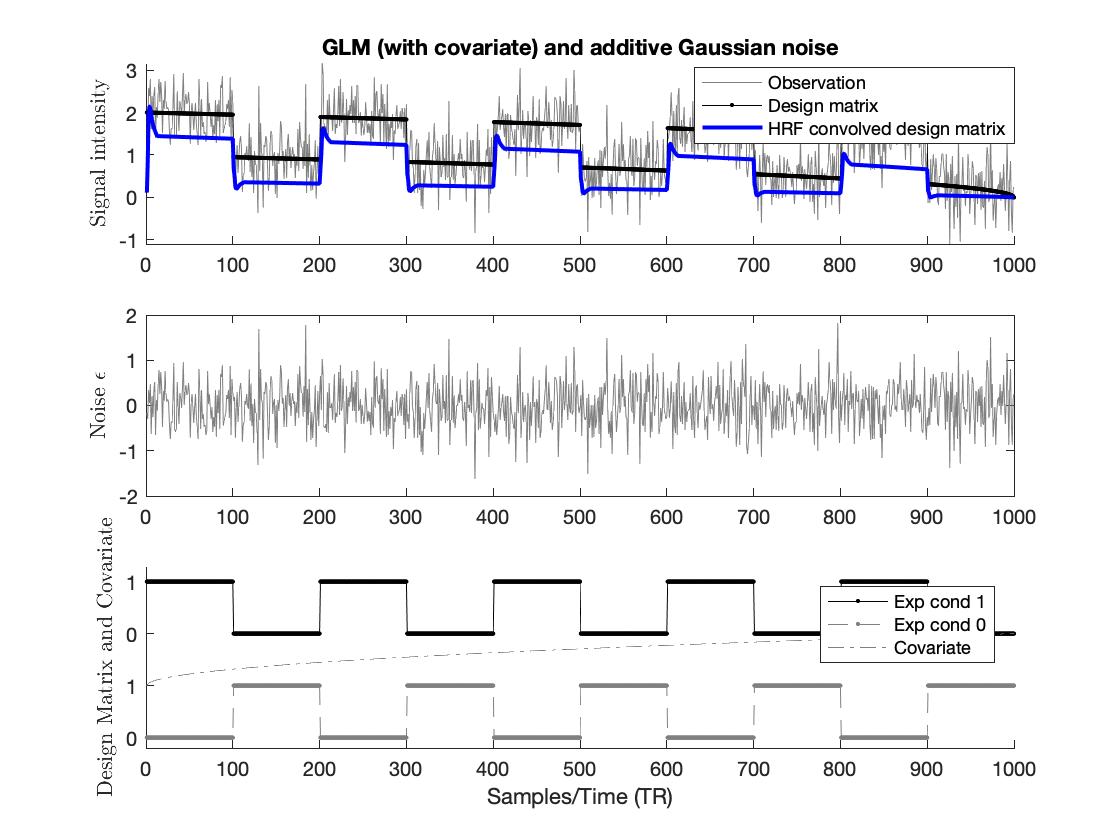}
\caption{Simulated data with noisy observations example with $\boldsymbol{\uptheta}=[1\quad 0\quad 1]^t$, CNR$=1$ and $N=1000$.}
\label{fig:uno}
\end{figure}

The  design matrix $\mathbf{X}$ was the canonical HRF convolved boxcar function for fMRI simulated data.  This matrix contained an exponentially decaying function $f(t)=(1-\frac{t}{N})^{0.5}$ representing  habituation during the fMRI task, or a covariate to simulate the effect of age when spatially testing  brain activations, and as shown in figure \ref{fig:uno}.  An $N$-dimensional Gaussian noise vector $\mathbf{v}$ was randomly drawn with zero mean and controlled variance ($\sigma_X^2/\text{CNR}$).  Finally,  a vector of observations was  constructed by adding the noise to the design matrix with ideal parameters $\theta=[1 \quad 0 \quad cv]^t$, where $cv$ is a constant that modulates the exponentially decreasing covariate in the time-series.

The use of synthetic data allowed us to estimated the noise covariance matrix by averaging a set of $100$ noise realizations as an ideal comparison with the GLM benchmark. This is useful to simulate different performances of the  ReML estimation that depends upon the observation vector $\mathbf{y}$ and some parametrization of the covariance components. The latter is the  procedure in SPM12 that obtains the noise covariance matrix $\mathbf{C}_{\epsilon}$ in equation \ref{eq:3}.


\subsection{A structural MRI dataset including covariates: ADNI}

Data were obtained from the Alzheimer's Disease Neuroimaging Initiative (ADNI) database (adni.loni.usc.edu). The ADNI database contains T1-weighted structural MRI scans acquired at 1.5 T and 3.0 T  from patients with Alzheimer's disease (AD), Mild Cognitive Impairment (MCI), and cognitively normal controls (NC)  at multiple time points. Here we only included structural MRI collected at 1.5T. The original database contains more than 1000 T1-weighted MRI images in total, although for this study only the first MRI examination of each participant was included, resulting in 417 structural images in the sample. Demographic data is summarized in Table \ref{tab:demog}.

The dataset was processed using the standard neuroimaging methods and protocols implemented in the SPM software (www.fil.ion.ucl.ac.uk/spm/), including registration to MNI space by spatial normalization and segmentation, to generate maps of grey matter (GM) volume \cite{Friston95}. 

Following the recommendation of the National Institute on Aging and the Alzheimer's Association for the use of imaging biomarkers \cite{NIA18}, we considered the group comparison NC vs.  AD for establishing a clear framework for comparing statistical paradigms. Age, sex and intracraneal volume (ICV) were included as covariates in between-group modelling as representing the most common set in the extant literature \cite{Hyatt20}. All the covariates were standardized with zero mean and standard deviation equal to one.

We selected two regions of interest using the 116-area automated anatomical labeling template \cite{Tzourio02}: one relevant area in AD, the bilateral hippocampus (denoted $Hippocampus_L$ and $Hippocampus_R$ for left and right hemispheres, respectively)  with 2559 voxels; and the cerebellum ($Cerebelum9_L$ and $Cerebelum9_R$ of the atlas), which is considered not relevant to AD pathology, with 3027 voxels. 

\begin{table}[htbp]
   \centering
   \caption{Demographic details of the MRI ADNI dataset, with group means and their standard deviations}
   \label{tab:demog}
   \begin{tabular}{lcccccc}
     \hline
     & Status & N &	Age	& Sex (M/F) &	ICV($\times 10^5)$ & MMSE\\ \midrule
& NC	        &   229    &	  75.9$\pm$5.0	    &   119/110	 & 15.3$\pm$ 1.6 & 29.1$\pm$1.0 \\
& AD	        &  188	   &     75.3$\pm$7.5 	&     99/89	& 15.5$\pm$ 1.8 & 23.2$\pm$2.2  \\ \bottomrule
\end{tabular}
\end{table}

\subsection{Statistical Analyses}\label{sec:inference}

To assess the performance of the methods presented in this paper, we used the benchmark proposed in SPM12 for statistical inference; that is, the GLM and RFT with FWE correction and $p=0.05$ for second-level statistical inference. 

First, we estimated the best set of parameters by regressing the design matrix, or the observations, using several configurations: i) the ideal ML method with synthetic data, where $\mathbf{C}_{\epsilon}$ is estimated from noise realizations, ii) Restricted (Re)ML, iii) LS, and iv) SVR. Then, we connected the MLE-based estimates (LS and SVR) with the corresponding set of parameters in the GLM space, as shown in equation \ref{eq:15}. To compare the inferences of each regression method, we assumed the same noise model and evaluated the T statistic in equation \ref{eq:3} on the set of parameters. Finally, we thresholded the resulting T-maps, e.g. derived from the SVR-iGLM method, by a detection threshold based on RFT.  

A permutation analysis was also adopted to provide an alternative statistical inference based on a non-parametric approach \cite{Mouro-Miranda05}. By randomly permuting the experimental conditions 1000 times, we calculated the non-parametric T statistic  based on the contrast of the estimates, thus avoiding the estimation of the denominator of equation \ref{eq:3}).The probability of observed contrast was then calculated relative to the distribution of permuted contrasts representing the null distribution. If this value was less than a selected threshold, e.g. $p=0.05$, then we rejected the null hypothesis.

\section{Experimental results}

\subsection{Synthetic data: Estimating functional tasks}

With the synthetic data we estimated the parameters by all the methods followed by permutation inference. We additionally plotted the non-normalized statistic (contrast) and evaluated the set of parameters, as a classifier, in the label domain, $\boldsymbol{w}\cdot \mathbf{y} \begin{array}{c} >\\ <\end{array} 0$. 

\begin{figure}
\centering
\includegraphics[width=0.5\textwidth]{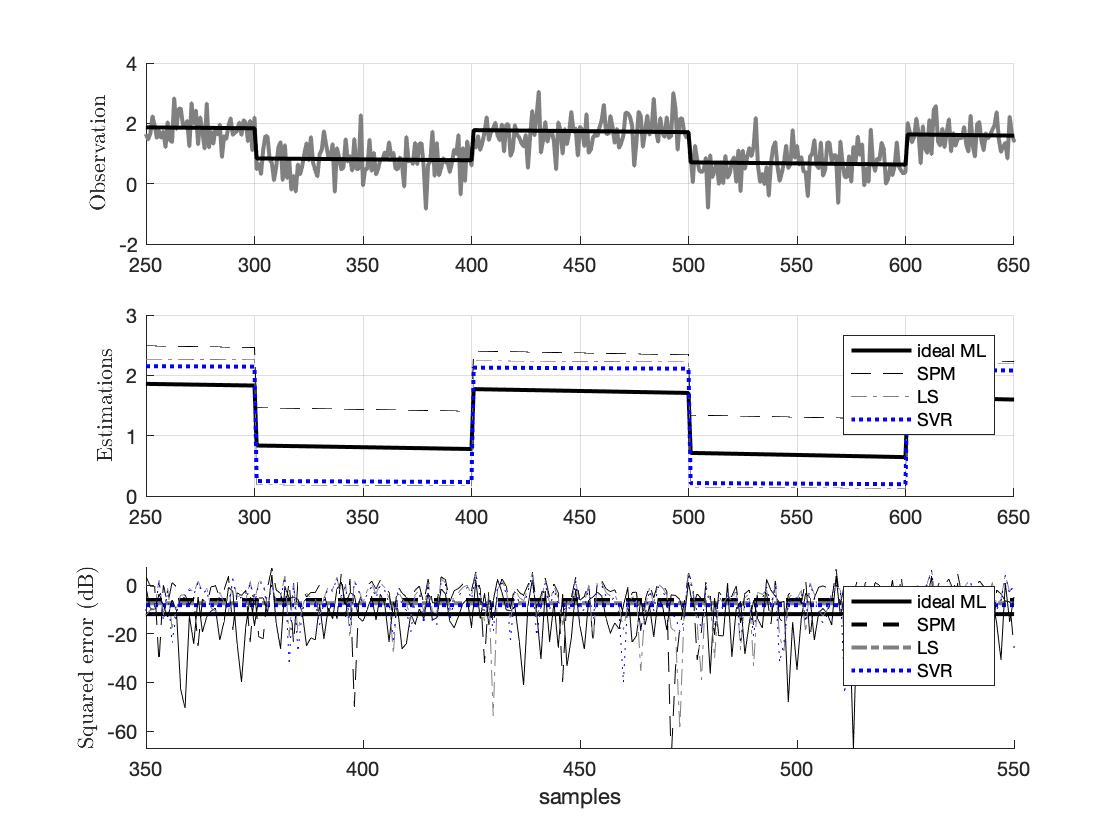}
\includegraphics[width=0.5\textwidth]{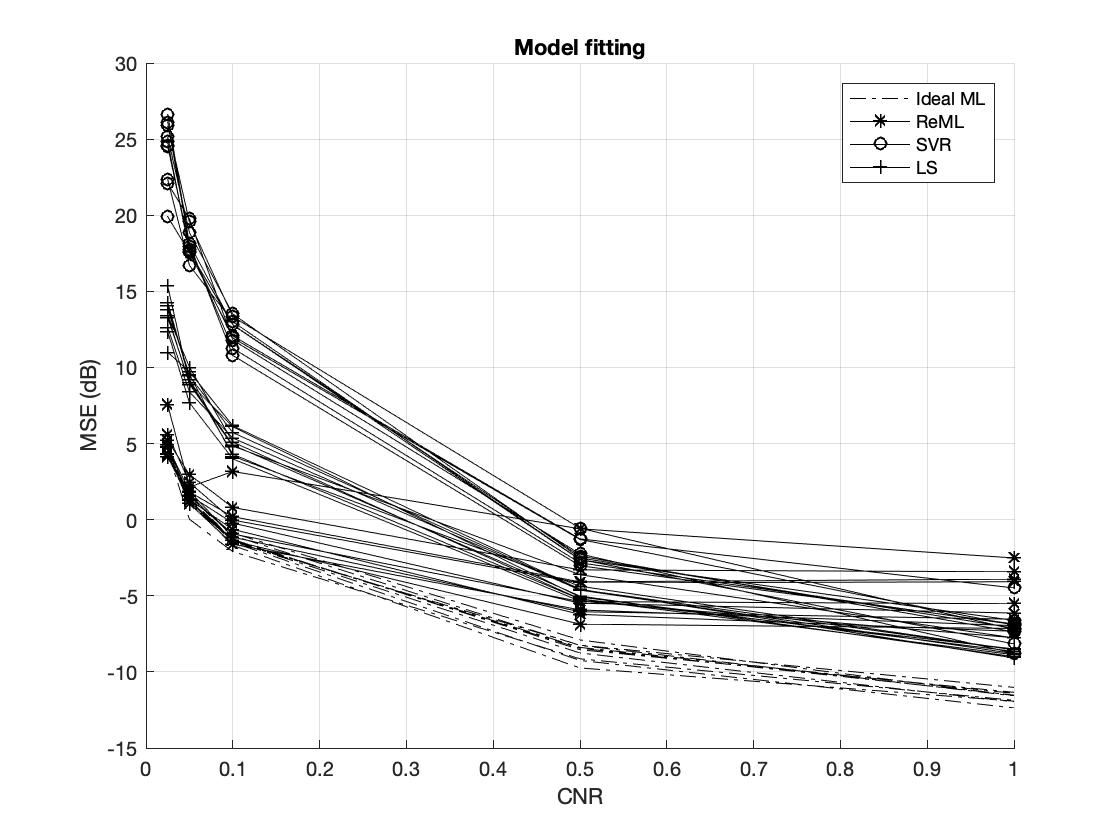}
\caption{Top: Example estimation with $N=1000$, CNR$=1$. Bottom: Averaged mean squared error (MSE) with different activation CNR and sample sizes.}
\label{fig:dos}
\end{figure}

\begin{figure}
\centering
\includegraphics[width=0.5\textwidth]{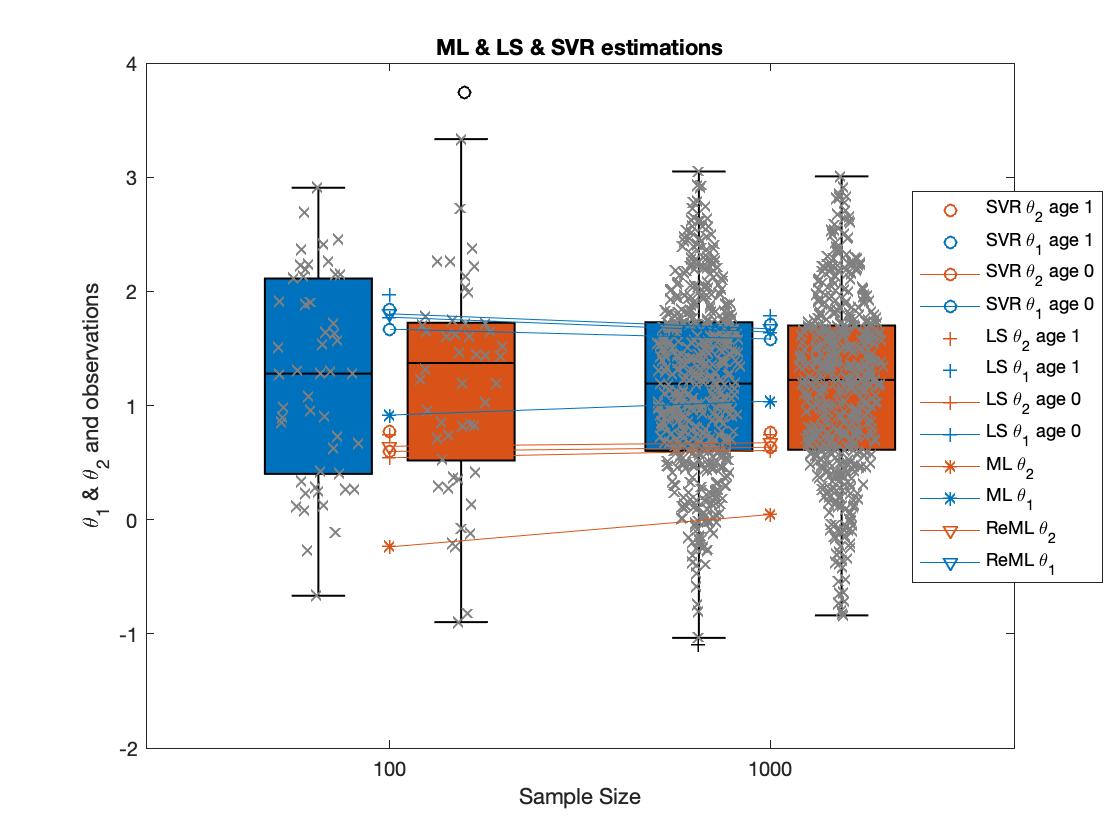}
\caption{Distribution of observations and estimations of $\boldsymbol{\uptheta}$  for all the analysed methods including the covariate effect.}
\label{fig:tres}
\end{figure}

The observations were artificially drawn from the same Gaussian sample distribution (figure \ref{fig:uno}) with varying sample size (i.e. time points) and CNR. We regressed both explanatory variables and observations to obtain the experimental parameters for each model $\boldsymbol{\uptheta}$, $\mathbf{w}$ given the simulated covariate. All these estimations were employed to calculate the regressed observed variables using equations \ref{eq:1} and \ref{eq:10}, given the explanatory matrix and the estimated parameters, and illustrated in figures \ref{fig:dos} and \ref{fig:tres}. Only for extremely noisy observations (CNR$<0.5$) and considerably high sample sizes did the ReML (and the ideal ML) outperform MLE approaches in MSE, even under the Gaussian assumption (figure \ref{fig:dos}). However, in the label domain, the set of parameters derived from the MLE approaches provided the highest classification accuracy in the associated classification task, recapitulating prior findings \cite{Gorriz2021}. In summary, the ML parameter fitting in the GLM does not imply an accurate regression in the iGLM. 

\begin{figure}
\centering
\includegraphics[width=0.5\textwidth]{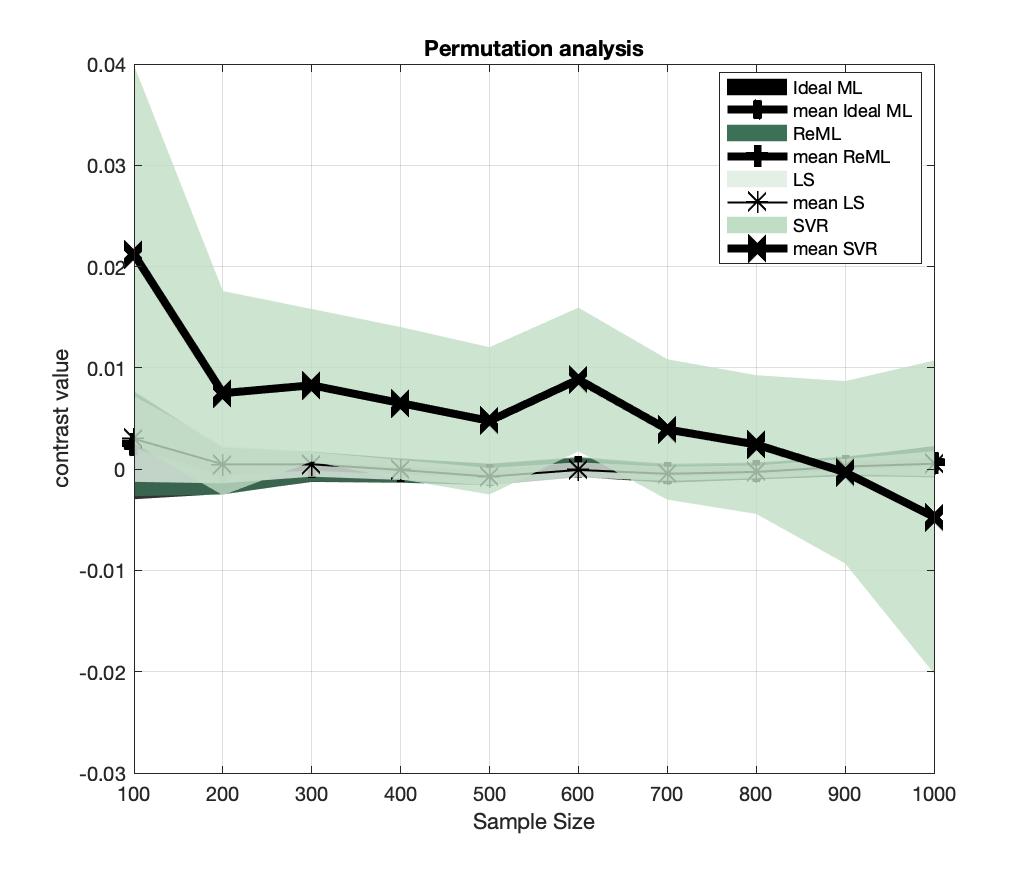}
\caption{Shadeplot showing contrast estimations under the null hypothesis for each method with its mean and the standard error, calculated from 1000 random permutations. CNR$=1$.}
\label{fig:cuatro}
\end{figure}

The p-value of the estimated contrast (difference between parameters $\theta_1-\theta_2$) was less than $p=0.05$ for all methods. The probability of observation was $p=1/1001$ in all cases when testing the null hypothesis. Thus, no false positive were detected during the simulated task, although the SVR estimation based inference provided results closest to the nominal false positive rate, whilst the remainder were  over-conservative  (figure \ref{fig:cuatro}).

\subsection{Empirical data: a case-control design with the ADNI Dataset}

In this section we show the inference derived from the two methodologies in each domain. We regressed on the observations and on the design matrix including covariates for age, sex and ICV (table \ref{tab:demog}). Then, we constructed the spatially extended statistical processes, generating maps of significance, using GM estimated from the MRI ADNI dataset\cite{Gorriz2021}. We compared the SPM (a two-sample T-statistic similar to equation \ref{eq:3}), where significance is first individually assessed at each voxel, and then on clusters with $p = 0.05$ FWE corrected based on RFT. The univariate test based on the inverse GLM in section \ref{sec:GLMleast} was also conducted.

\subsubsection{Hippocampal analysis to assess statistical power}

First, we performed the univariate statistical analysis on GM in the left and right hippocamppal ROIs, which are typically the earliest sites of atrophy associated with AD \cite{Johnson12}. Thus, the group comparison (advanced stage of the disease) AD vs. NC can be considered as a true positive (TP) region. We compared the GLM contrast with that obtained by the MLE-based approach using the same parametrization of the noise covariance in equation \ref{eq:3}, the standard approach in SPM12. SPMs were collected voxelwise and the inference method, based on a two-sample T-test with the linear compound $c=[1,-1,0,0,0]$, was thresholded by means of the RFT FWE rate correction at $p=0.05$, as previously.

\begin{figure}
\centering
\includegraphics[width=0.5\textwidth]{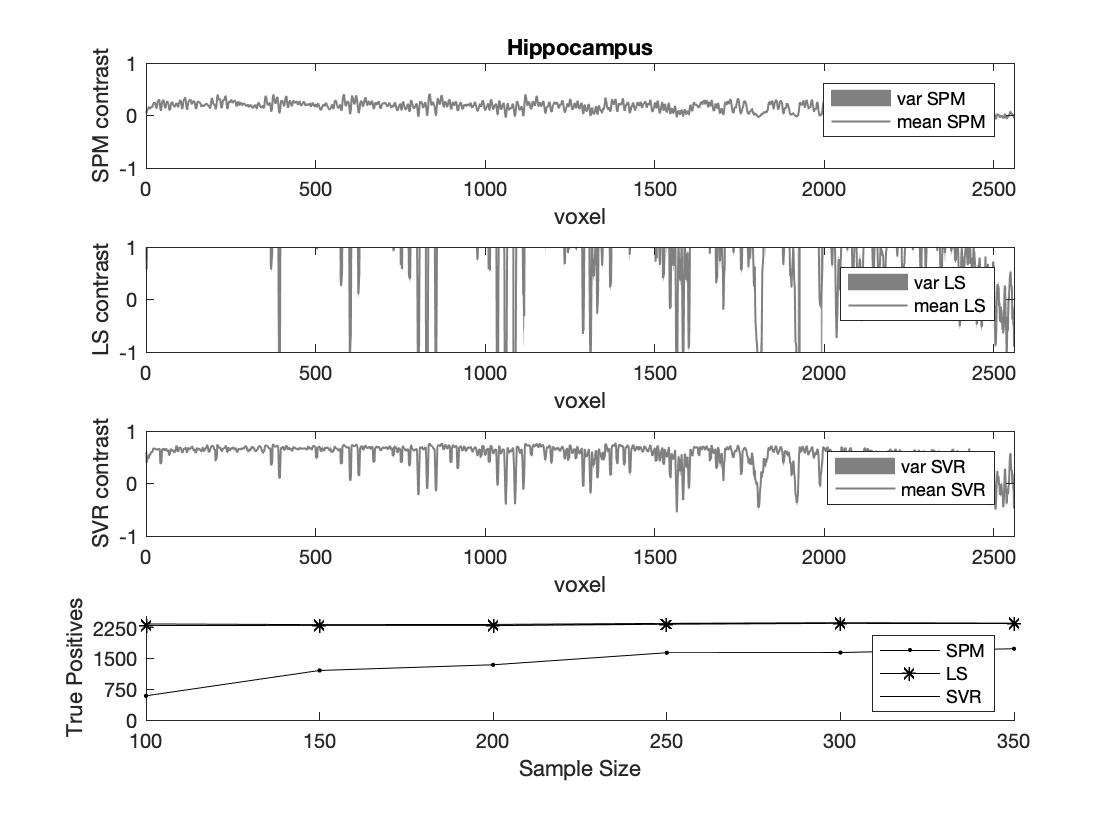}
\includegraphics[width=0.5\textwidth]{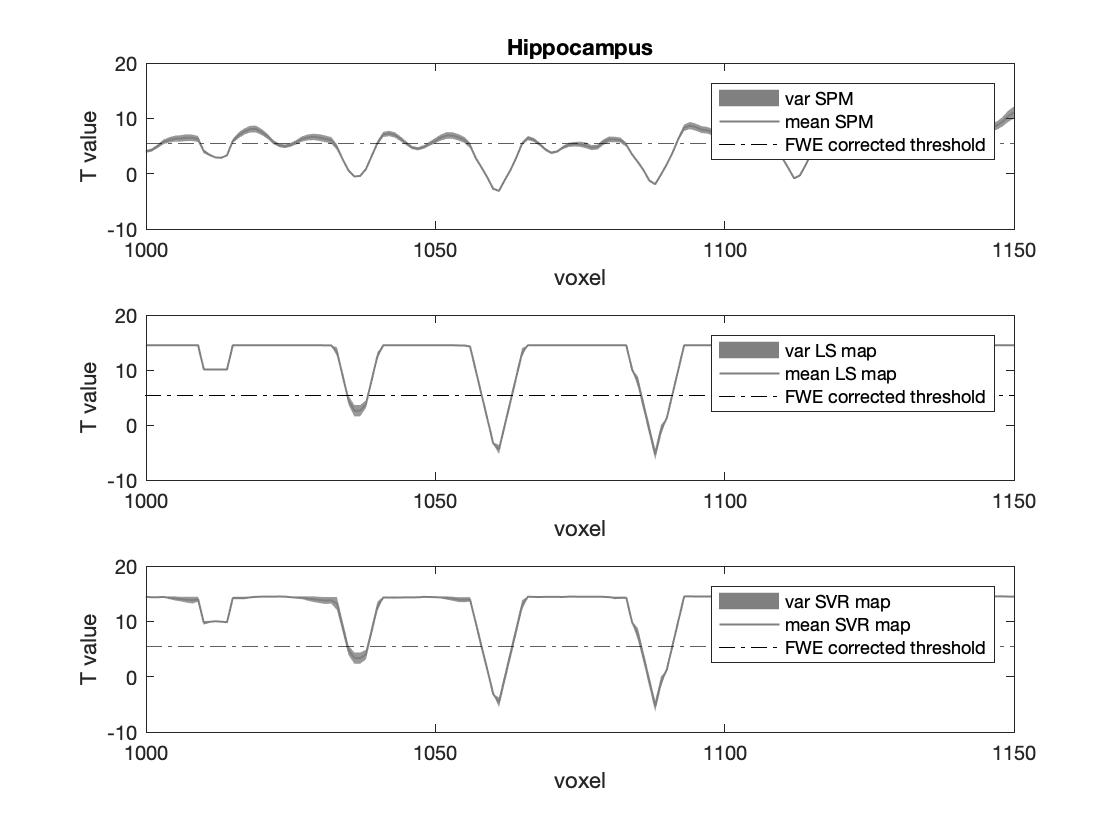}
\caption{Contrasts and T-statistics in the hippocampus with RFT FWE-corrected threshold at $p=0.05$ and varying sample sizes. Note that the MLE approaches are upper bounded for visualization purposes and the threshold plotted $T_{thres}=5.2758$ is that obtained for $N=350$.}
\label{fig:cinco}
\end{figure}

\begin{figure}
\centering
\includegraphics[width=0.25\textwidth]{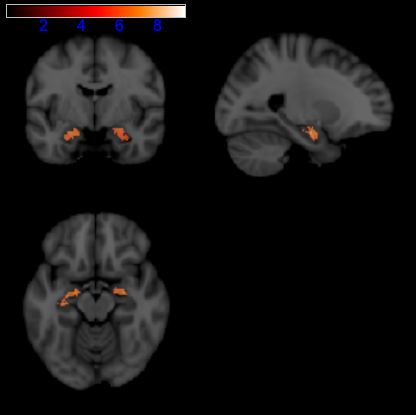}
\includegraphics[width=0.25\textwidth]{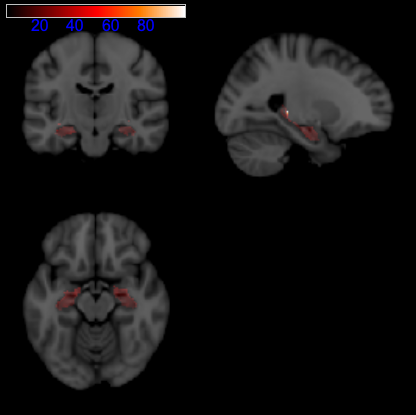}
\caption{Group-level statistical analysis results in the hippocampus of the ADNI MRI data. The univariate analysis was conducted using the standard GLM (top) and SVR-RFT approach (bottom). The T-maps were thresholded at $u>5.5704$ (equivalent to an uncorrected $p<1.1802e^{-7}$) with sample size $N=100$.}
\label{fig:cincobis}
\end{figure}

Figures \ref{fig:cinco} and \ref{fig:cincobis} show the contrasts (shadeplots with increasing sample size) derived after model estimation, and the T-statistics obtained by the use of the same normalization term in equation \ref{eq:3}. In this simulation, we employed the T-scores obtained by the standard SPM approach to bound the values obtained by our method. We readily see the higher statistical power ($T>T_{thres}=5.2758$ for $N=350$) of the MLE methods and the strong dependence of the standard SPM on the sample size. Both methods converged as sample size increased. The LS approach was more unstable with fewer significant voxels than the SVR-based procedure ($N=100$, $2314 vs 2341$; $N=150$, $2300 vs. 2324$;  $N=200$, $2300 vs. 2329$; $N=250$, $2334 vs. 2355$; $N=300$, $2355 vs. 2372$); $N=350$, $2344 vs. 2360$).

\subsubsection{Type I error control in a putatively-null cerebellar region}

\begin{figure}
\centering
\includegraphics[width=0.5\textwidth]{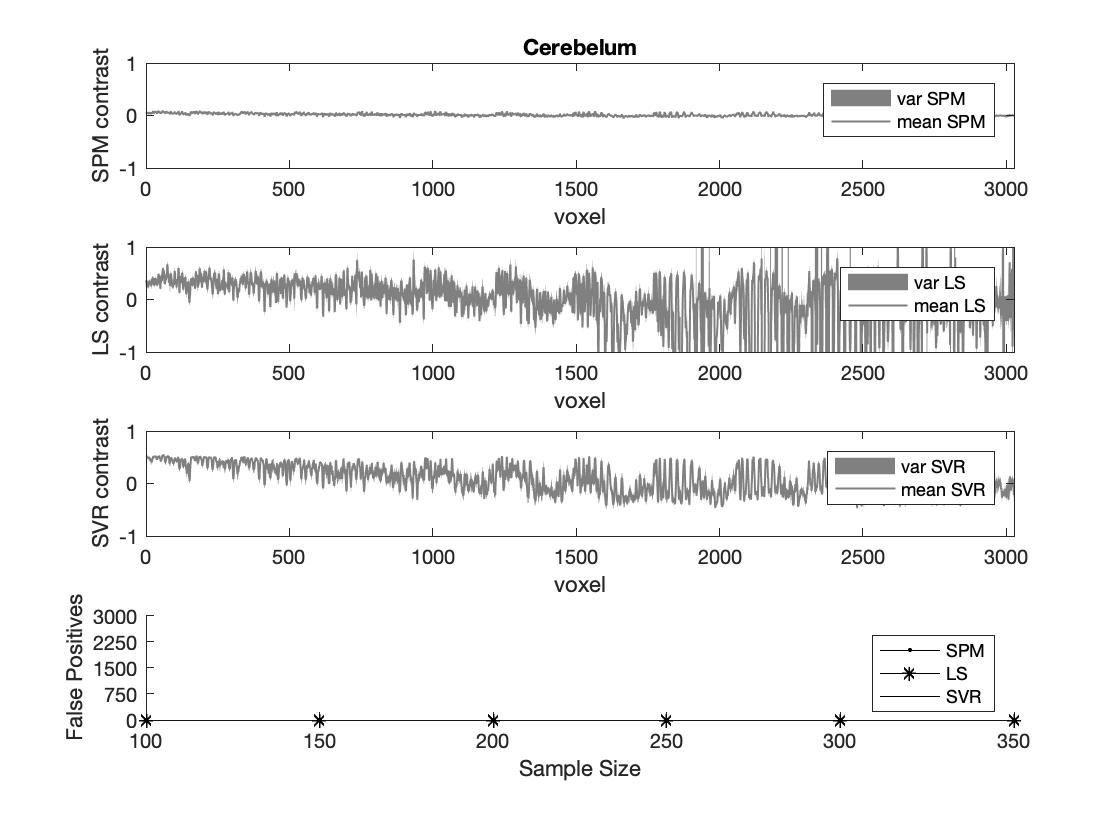}
\includegraphics[width=0.5\textwidth]{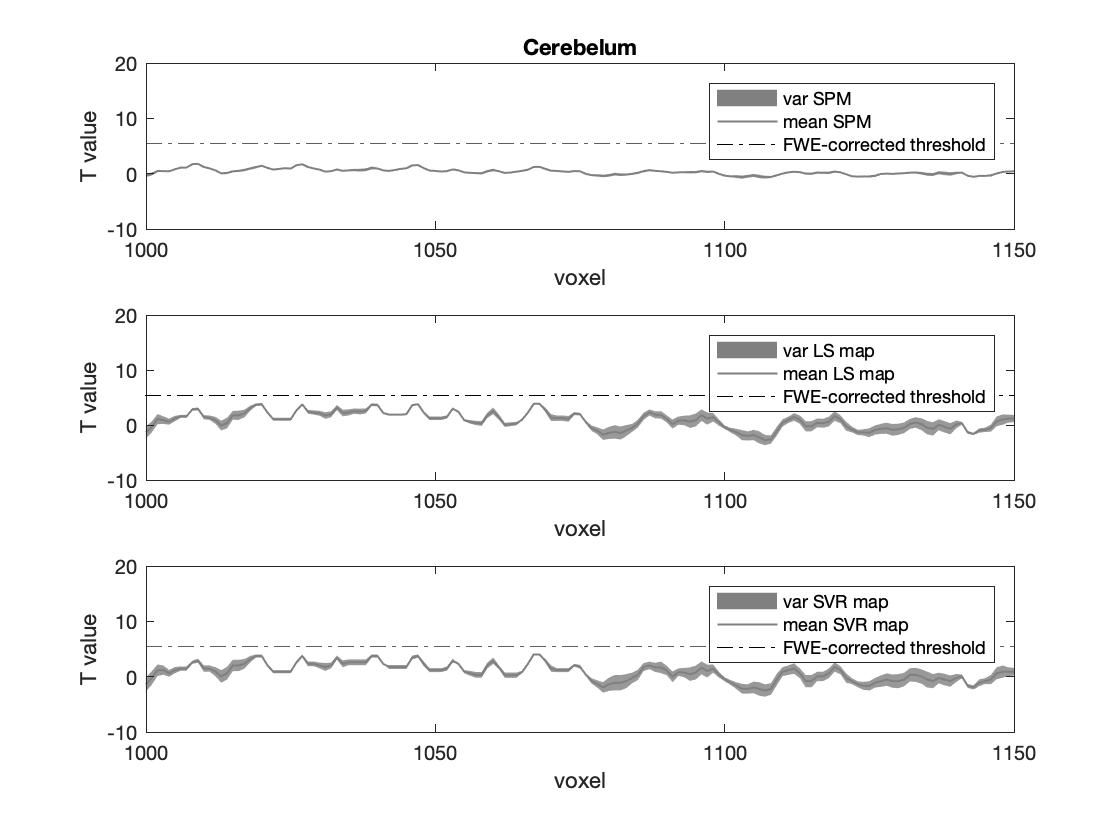}
\caption{Contrasts and T-statistics for the cerebellum with FWE-corrected threshold $t_{thres}=5.2758$ at $p=0.05$.}
\label{fig:seis}
\end{figure}

We repeated the experiments on the GM in cerebellar ROIs (Cerebellum $9$ left and right \cite{Tzourio02}) to evaluate the ability of the MLE-based inference methods to control false positive rates using the same inference strategy used for thresholding the T-maps; i.e. FWE correction based on RFT in the standard GLM. Figure \ref{fig:seis} illustrates the over-conservative voxelwise inference with FWE correction based on RFT, although the proposed methods based on MLE provided a significant number of tests closer to the expected value, i.e. $p=0.05$. In other words, by chance the number of FPs should be around $5\%$ of the total number of voxels. Parametric voxelwise inference is known to be valid but conservative, often falling below the nominal rate \cite{Eklund16}.

\subsubsection{Global analysis in contrast images with limited sample size}

\begin{figure}
\centering
\includegraphics[width=0.5\textwidth]{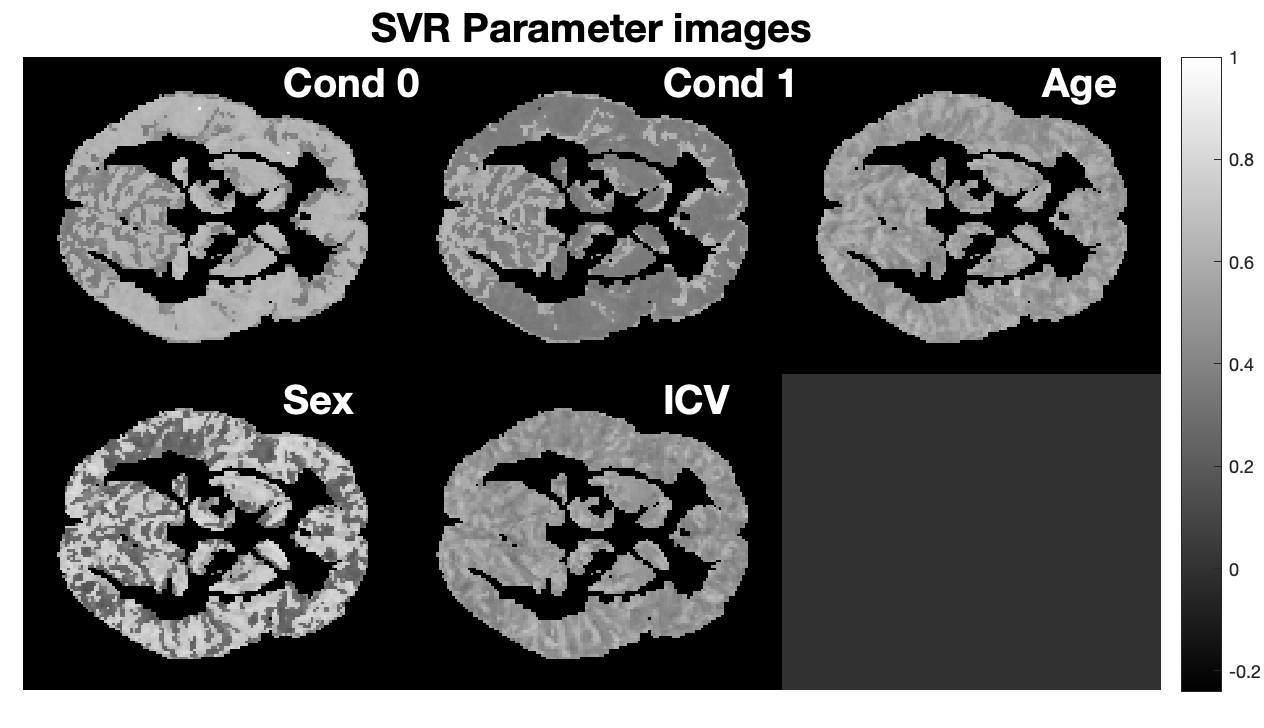}
\includegraphics[width=0.5\textwidth]{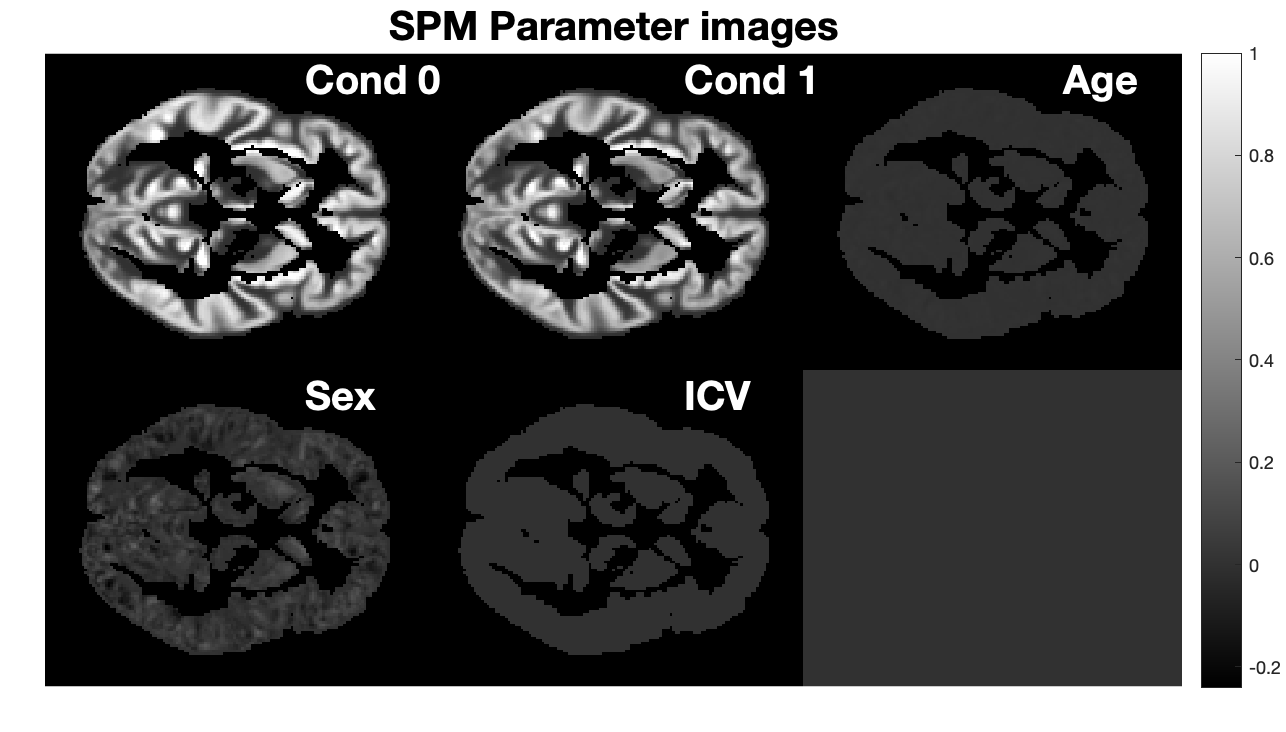}
\caption{Parameter images (axial slice Z=$47$) derived from ReML (SPM) and SVR estimations for $N=100$. Note that Cond 0 and Cond 1 denote NC and AD, respectively.}
\label{fig:siete}
\end{figure}

Figure \ref{fig:siete} shows the vector of parameters in the selected axial slice Z=$47$ derived from ReML and SVR in the analysis of the whole brain. The ML approach implemented in SPM  finds very few significant relationships between the covariate effects (age, sex and ICV) and the observations, compared with the experimental conditions. On the contrary, the SVR yields stronger connections, mainly for the sex covariate. From these parameter images, contrast images are then derived and, subsequently,  inference on their size relative to the estimate of their standard error is made in large target regions, as depicted in figures \ref{fig:cinco} and \ref{fig:seis}. 

\begin{figure}
\centering
\includegraphics[width=0.5\textwidth]{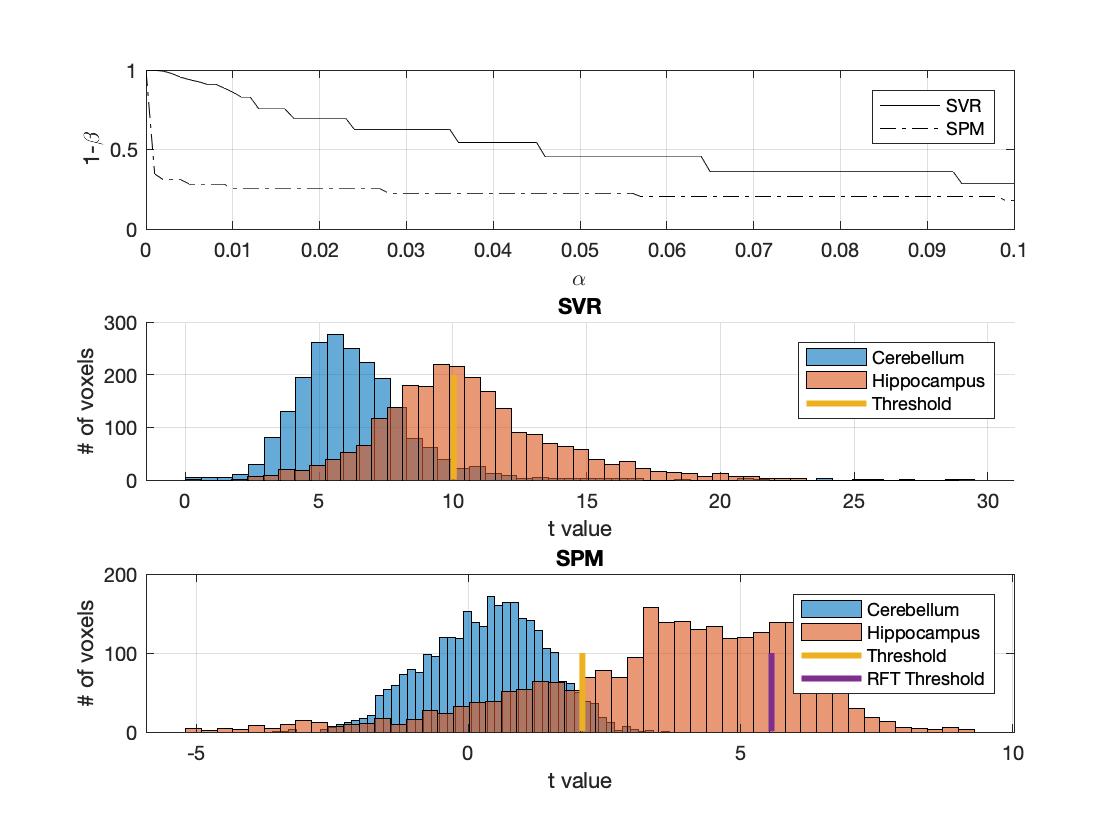}
\caption{Power analysis using the optimum Neyman-Pearson threshold. We selected a level of significance of $\alpha=0.05$ that resulted in a threshold less conservative than the RFT correction.}
\label{fig:ocho}
\end{figure}

\begin{figure*}
\centering
\includegraphics[width=0.5\textwidth]{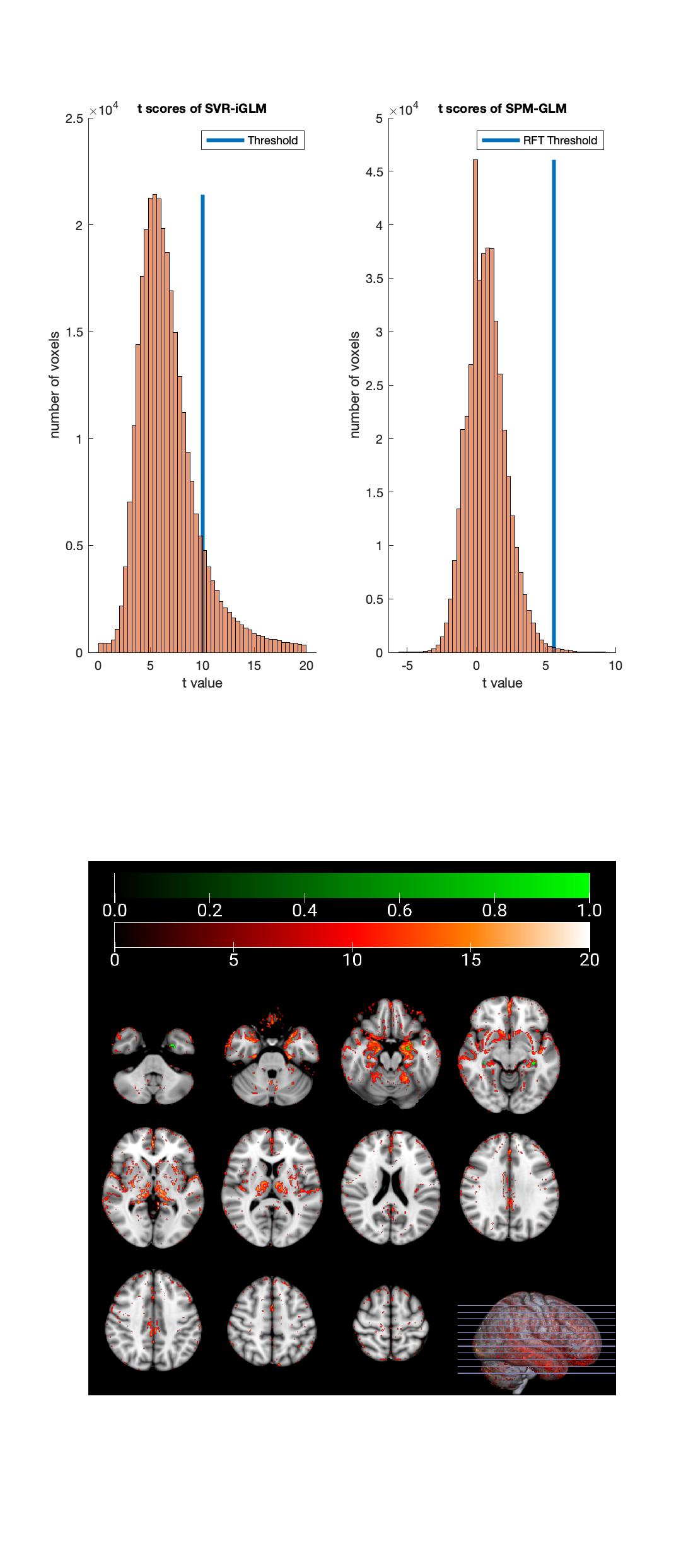}
\caption{Global T-scores and activation maps of the whole image volume, obtained by the Neyman-Pearson threshold at $\alpha=0.05$ and the standard SPM FWE $p=0.05$ correction.}
\label{fig:nueve}
\end{figure*}

Selecting the aforementioned target regions and plotting the t-score histograms we can approximately compute the optimum threshold at a given level of significance $\alpha$ using the Neyman-Pearson lemma, e.g. at $\alpha=0.05$, SVR: $t_{thresh}=10.03$ and SPM: $T_{thresh}=2.03$. By comparing the latter value with the one used in previous sections (RFT correction) we can readily see the over-conservative nature of the standard voxelwise inference (figure \ref{fig:ocho}). Extending these thresholds to the whole volume we obtain the activation maps for the SVR and standard SPM approaches as shown in figure \ref{fig:nueve}.

\section{Discussion}

In the context of neuroimaging statistical inference, there is an increasing trend to incorporate exploratory methods into well-established GLM-based data analysis. Not only data preprocessing techniques, such as independent or principal component analysis (PCA) \cite{McKeown03}, but also multivariate MLE approaches have been widely used in classification tasks to replace the predefined design matrix in the regular GLM pipeline \cite{Wang09} or to provide novel statistical maps of prevalence \cite{Mouro-Miranda05,Gorriz2021}. In this paper we present a novel univariate methodology for (f)MRI image analysis based on the optimum performance in limited sample sizes of SVR over classical approaches. It departs from the use of the typical GLM frameworks based on classical estimations, or hybrid approaches that combine GLM with MLE, and proposes a complete voxelwise inference method based on SVR. This methodology was demonstrated to provide optimum generalization ability in the case of regression estimation by the construction of regularizing functions in ill-posed problems \cite{Vapnik82}. 

The inverse GLM-based method (SVR-iGLM) is an univariate approach that preserves the advantages of classical approaches, such as function localization and better interpretation, and incorporates the aforementioned advantages. Existing multivariate approaches based on MLE, including stages for smoothing or orthogonal decomposition such as PCA \cite{Mouro-Miranda05,Wang09,Wang07,Gorriz2021}, have provided promising results in (f)MRI-data analysis where there is a trade-off between sensitivity and computational cost. We explored the performance of the univariate MLE approach in the inverse GLM and compared it with the regular GLM inference. As in previous approaches we demonstrated the highest detection ability where its sensitivity was controlled with a common p-value correction in the standard GLM and with the optimum Neyman-Pearson threshold in the whole volume analysis.

Although covarying for data variables in neuroimaging is routine when adjusting the model for confounding or nuisance factors \cite{Hyatt20}, data-driven approaches have limited their operation to extracting whole brain SDMs as a description of the different responses to experimental conditions.  Additional covariates can be avoided in MLE approaches by the selection of balanced groups, sometimes a complex task, with the aim of reducing the impact of predictors that are not relevant to the research question at hand \cite{Leming20}. However, SDM extraction is not possible for multiple conditions/states in fMRI data due to the use of the signed distance from the hyperplane \cite{Wang09} in a classification task. Therefore, group-balanced selection procedures further decrease sample sizes. The proposed MLE approach, linked to the regular GLM, processes all the covariates at once, and combines their effect when estimating the observation or response variable. This effect is quantitatively determined by the definition of the equivalent vector of parameters $\tilde{\boldsymbol{\uptheta}}$.  We tested the SVR estimation within a permutation inference and the regular GLM framework with FWE correction based on RFT. With synthetic data, the SVR approach resulted in a similar mean squared error as the GLM based on ReML estimations for CNR$>0.5$, whilst with real data the MLE contrasts showed greater variability between conditions. The univariate SVR approach had improved discrimination compared to the GLM based inference, aligning with previous results in the extant literature that employed multivariate MLE approaches to classify experimental conditions under the GLM. The proposed approach provides: i) inference results closer to the nominal values, i.e. $p=0.05$, although over-conservative behaviour prevails; and ii) robust performance with increasing sample size.

Several limitations are known using LS linear regression models (LRM) for estimating the vector of parameters \cite{Gorriz2021,Hastie2001}, and many more due to instabilities of the algorithm, including effects of outliers, heteroskedasticity, etc. However, as shown in the experimental section, there was a strong correspondence between GLM and LS-LRM in the synthetic data analysis. Unfortunately, univariate MLE approaches increase the computational burden of the analysis since they perform multiple LRMs independently at each voxel in the image. This is a drawback for fMRI analyses where the number of scans and voxels is inflated relative to structural images. Nevertheless, this issue is usually  solved by the use of spatial dimension reduction and data representation techniques, such as Partial Least Squares (PLS) or PCA \cite{Gorriz18,Wang07} in multivariate frameworks.  Another common caveat of the univariate MLE approaches in this scenario is overfitting \cite{Scholkopf02}, a well-known problem in pattern recognition. Here, the regressor, once fitted, conforms to the specific samples in the training data with the consequence that its generalization ability or sensitivity to unseen data is reduced. The selected SVR employs the concept of maximizing the margin with a regularized term and a low-complexity model to reduce the risk of overfitting. Nevertheless, the fitting process could be affected by misregistration \cite{Rosenblatt14} or spatially incoherent activations \cite{Wang07}. The use of smoothing kernels in univariate approaches, such as SVR-iGLM, could potentially reduce this effect by taking spatial correlations into account, but could conversely worsen sensitivity in incoherent signals in multivariate approaches.

Most hypothesis-driven neuroimaging analyses depend on specified models when proposing a statistic and fitting parameters of the GLM. This is a major advantage when data (nature’s mechanism) is drawn assuming a Gaussian distribution (model’s mechanism), and the inference drawn from the experiment may be misleading \cite{Breiman01}. In the synthetic example, we assumed a known covariance matrix and a set of noise realizations for the formulation of the GLM. This experimental setup is imperfect in neuroimaging applications and the statistics following on from the best guess can fluctuate around the ideal value \cite{Gorriz2021}. Frequentist and Bayesian analyses are strongly grounded on model selection and parameter fitting stages where in complex scenarios with a limited sample size heuristics are the common solution \cite{Woolrich2009}.  Finally, limited samples sizes and the selection/estimation of any specific model remains an issue in neuroimaging. This problem potentially deteriorates if the model, and the interaction between model parameters, becomes too complex for an accurate posterior probability estimation or a feasible numerical computation of Bayes rule. Given the relationship between the GLM and MLE-based regression, we propose a conventional statistical inference based on the optimum estimations derived from MLE  from limited amounts of data  \cite{Vapnik82,Haussler92}. The SVR-iGLM is not limited to linear regression since the main regressor in the design matrix could be replaced by another non-linear function; a common approach in fMRI data modelling. Moreover, SVR-iGLM could be incorporated in novel statistical tests, e.g. the P-tests \cite{Reiss15}, to highlight between-group differences in patterns of imaging-derived measures. 

\section{Conclusions}

We addressed the open question on the usefulness and the interpretation of MLE approaches for obtaining spatial patterns from brain imaging data that can discriminate between samples or brain states. We followed the natural path for using the MLE framework by regressing observations onto conditions in a supervised learning manner, including a set of covariates. We thus explored the complete connection between the univariate GLM and MLE \emph{regressions} by deriving a refined statistical test based on the parameters obtained by a SVR in the \emph{inverse} problem. Experimental results demonstrated how parameter estimations derived from MLE and common statistical inference procedures provide a novel technique with good statistical power and control of false positives to obtain SDMs.


\newpage


\section{Supplementary Material}
\label{Appendix1}

For simplicity, we restrict the discussion to the special case of having two explanatory variables ($M = 2$). In this case,~\eqref{eq:1} reads:
\begin{equation}
\label{eq:GLM-two-covariates}
{\mathbf{y}} = \mathbf{x}_{:,1} \, \theta_{1} + \mathbf{x}_{:,2} \, \theta_{2} + \boldsymbol{\upepsilon}.
\end{equation}
where $\mathbf{x}_{:,i}$ represents here the $i$th column of the indicator matrix ${\mathbf{X}}$. As this matrix codifies the membership of the observations, $x_{i,j}$ is either zero or one, where a one indicates that $y_{i}$ is grouped into the $j$-th class. Supposing that the classes are disjoint, the indicator matrix can have only a single one per row. Therefore:
\begin{equation}
\label{eq:disjoint-classes}
\mathbf{x}_{:,1} + \mathbf{x}_{:,2} = \mathbf{1},
\end{equation}
where $\mathbf{1}$ is the $N$ vector of ones.  Using~\eqref{eq:disjoint-classes},~\eqref{eq:GLM-two-covariates} can be rewritten as:
\begin{equation}
\label{eq:GLM-simplified}
{\mathbf{y}} =   \mathbf{x}_{:,1} \, (\theta_{1} - \theta_{2}) + \mathbf{1} \, \theta_{2}  + \boldsymbol{\upepsilon}.
\end{equation}
For mathematical convenience, we subtract out the mean of the observations by premultiplyng~\eqref{eq:GLM-simplified} by matrix $\mathbf{P} = \mathbf{I} - \frac{\mathbf{1}\mathbf{1}^t}{N}$, where $\mathbf{I}$ is the $N\times N$ identity matrix. By so doing, we get the centered model
\begin{equation}
\label{eq:GLM-centered}
\vec{\mathbf{y}} =  \vec{\mathbf{x}}_{:,1} \, (\theta_{1} - \theta_{2}) +  \vec{\boldsymbol{\upepsilon}},
\end{equation}
where
\begin{equation*}
\vec{\mathbf{y}} = \mathbf{P}\,{\mathbf{y}}, \quad \vec{\mathbf{x}}_{:,1} = \mathbf{P}\,\mathbf{x}_{:,1} \text{ and } \vec{\boldsymbol{\upepsilon}} = \mathbf{P}\,\boldsymbol{\upepsilon}.
\end{equation*}
By construction, all of them are zero-mean vectors. Among other things, this implies that:
\begin{remark}
\label{th:remark1}
Since $\mathbf{x}_{:,1}$ is a vector of zeros and ones, both with the same probabilities, $\vec{\mathbf{x}}_{:,1}$ has entries $1/2$ or $-1/2$, where the sign indicates the class membership.
\end{remark}

Let us now consider the inverse model
\begin{equation}
\vec{\mathbf{x}}_{:,1} = \vec{\mathbf{y}} \, \omega + \hat{\boldsymbol{\upepsilon}}
\end{equation}
and solve for $\omega$ using two different approaches. Note that, since every vector is of zero mean, the inverse model does not require a bias term.

\subsection{MSE approach}

We first consider the classical minimum squared error (MSE) estimator, which will serve as the gold standard for the subsequent approaches. MSE consists in minimizing
\begin{equation}
\label{eq:MSEcriterion}
\| \hat{\boldsymbol{\upepsilon}} \|^{2} = \|\vec{\mathbf{x}}_{:,1} - \vec{\mathbf{y}} \, w\|^{2},
\end{equation}
where $\|\cdot\|$ denotes the Euclidean or $L_{2}$ norm. Simple calculus shows that the minimum is attained at
\begin{equation}
\label{eq:mse-optimum}
\omega_\text{MSE} = \frac{\vec{\mathbf{y}}\cdot\vec{\mathbf{x}}_{:,1}}{\|\vec{\mathbf{y}}\|^{2}}.
\end{equation}
Consequently, $\vec{\mathbf{y}} \, \omega_\text{MSE}$ is simply the projection of $\vec{\mathbf{x}}_{:,1}$ onto the observation vector $\vec{\mathbf{y}}$. Now, by substituting~\eqref{eq:GLM-centered} in~\eqref{eq:mse-optimum}, we also get:
\begin{equation}
	\label{eq:MSEopt}
\omega_\text{MSE} = \frac{\theta_{1} - \theta_{2}}{(\theta_{1} - \theta_{2})^{2} + 4 \, \sigma_{\boldsymbol{\upepsilon}}^{2}},
\end{equation}
where $\sigma_{\boldsymbol{\upepsilon}}^{2} = \frac{1}{N} \|\vec{\boldsymbol{\upepsilon}}\|^{2}$ is the variance of the error term and we have supposed that $\vec{\mathbf{x}}_{:,1}\cdot\vec{\boldsymbol{\upepsilon}} = 0$. Observe that  $\omega_\text{MSE}$ approaches $1/(\theta_{1} - \theta_{2})$ as $\sigma_{\boldsymbol{\upepsilon}}^{2}\downarrow 0$, which shows that regression estimates  the parameters of the GLM.

\subsection{L1 norm-based approach} The $L_{2}$ norm is very sensitive to outliers. For this reason, $L_{1}$ norm approaches, which are more robust, are gaining in popularity. Indeed, SV regression is actually based on minimizing a $L_{1}$ norm-based criterion. For our problem at hand, we substitute the MSE cost~\eqref{eq:MSEcriterion} with
\begin{equation}
\label{eq:L1criterion}
\| \hat{\boldsymbol{\upepsilon}} \|_{1} = \|\vec{\mathbf{x}}_{:,1} - \vec{\mathbf{y}} \, \omega\|_{1}
\end{equation}
where $\|\cdot\|_{1}$ denotes the $L_{1}$ norm, that is,
\begin{equation}
\label{eq:L1-criterion}
\| \hat{\boldsymbol{\upepsilon}} \|_{1} = \sum_{n=1}^{N} | \hat\upepsilon_{n}|.
\end{equation}

As $\vec{\mathbf{y}}$ is given in~\eqref{eq:GLM-centered}, $\hat{\boldsymbol{\upepsilon}}$ can be recast as
\begin{equation}
\label{eq:epsilon}
- \hat{\boldsymbol{\upepsilon}} =  \omega \, \vec{\boldsymbol{\upepsilon}} + (v - 1) \, \vec{\mathbf{x}}_{:,1},
\end{equation}
where $v = (\theta_{1} - \theta_{2})\,w$. Consider the $n$-th component, $- \hat{\upepsilon}_{n} = \omega \, \vec{\upepsilon}_{n} + (v - 1) \, \vec{x}_{n,1}$, where $\vec{x}_{n,1}$ can be either $1/2$ or $-1/2$ with equal likelihood (recall Remark~\ref{th:remark1}). Suppose that $\vec{\upepsilon}_{n}$ is drawn from a $\mathcal{N}(0 \, , \, \sigma_{\boldsymbol{\upepsilon}}^{2})$. After shifting $w\,\vec{\upepsilon}_{n}$ by $(v - 1) \, \vec{x}_{n,1}$, with the application of Bayes' rule one sees that  $\hat{\upepsilon}_{n}$ comes from a population whose pdf, $p(\hat{\upepsilon})$, is the following mixture of Gaussians:
\begin{equation}
p(\hat{\upepsilon}) = \frac 1 2 \mathcal{N}(\hat{\upepsilon} \, | \, -\mu \,,\, \omega^{2} \, \sigma_{\boldsymbol{\upepsilon}}^{2}) + \frac 1 2 \mathcal{N}(\hat{\upepsilon} \, | \, \mu \,,\, \omega^{2} \, \sigma_{\boldsymbol{\upepsilon}}^{2}),
\end{equation}
where $\mu = \frac{(v - 1)}{2}$. This pdf enables us to derive a closed-form expression for the $L_{1}$ criterion~\eqref{eq:L1-criterion}, as follows: under the assumptions of the law of large numbers, $\frac{\| \hat{\boldsymbol{\upepsilon}} \|_{1}}{N}$ tends, as $N \uparrow \infty$, to
\begin{equation}
E\{|\hat\upepsilon|\} = \int_{-\infty}^{\infty} |\hat\upepsilon|\,p(\hat\upepsilon)\,\text{d} \hat\upepsilon.
\end{equation}

Then, using the identity $\int \beta \, e^{-\frac{\beta^2}{2\,\sigma^2}} \text{d} \beta = - \sigma^2 e^{-\frac{\beta^2}{2\,\sigma^2}} + \text{ constant}$, some algebra shows that
\begin{equation}
\label{eq:L1cost}
	E\{|\hat\varepsilon|\} = \frac{|\omega|}{\sqrt{2}}  \, \sigma_\varepsilon \, g\left(\frac{\mu}{\sqrt{2} |\omega| \sigma_\varepsilon}\right)
\end{equation}
where $g(\alpha) = \frac{e^{-\alpha^2}}{\sqrt{\pi}} + \alpha\,\text{erf}(\alpha)$ and $\text{erf}(\cdot)$ denotes the error function, defined as $\text{erf}(\alpha) = \frac{2}{\sqrt{\pi}} \int_{0}^\alpha e^{-\beta^2} \text{d}\beta$. The problem is to find the $\omega$ that minimizes~\eqref{eq:L1cost}. By taking the gradient of~\eqref{eq:L1cost} with respect to $\omega$ and equating to zero, we get the equation:
\begin{align}
\sqrt{\frac{2}{\pi}}\,\sigma_\upepsilon \, \exp\left(-\left[\frac{(\theta_1 - \theta_2) \omega - 1}{2 \sqrt{2} |\omega| \sigma_\upepsilon}\right]^2\right) & = \nonumber \\ 
\label{eq:optimum}
 - \frac{(\theta_1 - \theta_2)}{2 } \, \text{sign}(\omega) \, \text{erf}\left({\frac{(\theta_1 - \theta_2) \omega - 1}{2 \sqrt{2} |\omega| \sigma_\upepsilon}}\right). & 
\end{align}
Obviously, it cannot be solved analytically and, therefore, we resort to an approximation. Let us assume that close to the solution we have 
\begin{equation}
	\label{eq:assumpt}
	\frac{(\theta_1 - \theta_2) \, \omega - 1}{2 \sqrt{2} |\omega| \sigma_\upepsilon} \approx 0.
\end{equation}
Then, a first order Maclaurin series expansion of~\eqref{eq:optimum}, i.e., $e^{-\alpha^2} \approx 1$ and $\text{erf}(\alpha) \approx 2 \alpha / \sqrt{\pi}$, gives:
\[
\sqrt{\frac{2}{\pi}}\,\sigma_\upepsilon  \approx - (\theta_1 - \theta_2) \,  \frac{\text{sign}(\omega)}{\sqrt{\pi}} \frac{(\theta_1 - \theta_2) \, \omega - 1}{2 \sqrt{2} |\omega| \sigma_\upepsilon},
\]
and, solving for $\omega$, we get
\begin{equation}
	\label{eq:MSEopt}
	\omega^* \approx \frac{\theta_{1} - \theta_{2}}{(\theta_{1} - \theta_{2})^{2} + 4 \, \sigma_{\boldsymbol{\upepsilon}}^{2}},
\end{equation}
which coincides with the MSE solution~\eqref{eq:MSEopt}. It only remains to verify the assumption~\eqref{eq:assumpt}. As
\[
\frac{(\theta_1 - \theta_2) \, \omega^* - 1}{2 \sqrt{2} |\omega^*| \sigma_\upepsilon} = - \sqrt{2} \, \frac{\sigma_\upepsilon}{|\theta_1 - \theta_2|},
\]
we obtain that our approximation is acceptable as long as $\sigma_\upepsilon \ll |\theta_1 - \theta_2|$, which seems to be a reasonable hypothesis.

\subsection{Simulation}

\begin{figure}
	\centering
	\includegraphics[width=0.5\linewidth]{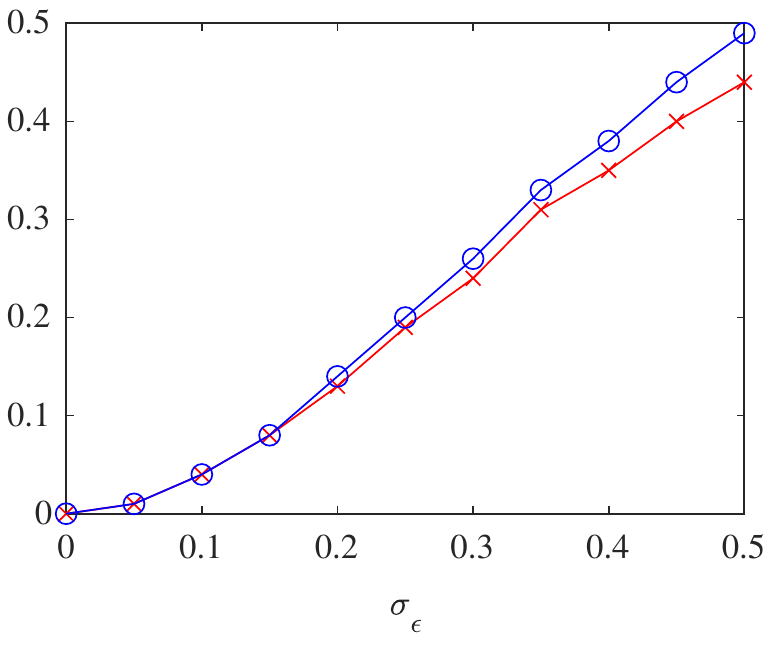}
	\caption{Absolute error committed in the approximation of the optimal value $w = 1/(\theta_1 - \theta_2)$, as a function of $\sigma_\varepsilon$, by using the L1-approach (in red) and the MSE approach (in blue).}
	\label{fig:error-criteria}
\end{figure}

To experimentally evaluate the performance of the above $L_1$ norm-based regression criterion, we draw a Gaussian random vector $\vec{\boldsymbol{\upepsilon}} \in {\mathbb R}^{N}$, with $N = 100$, where each entry $\vec\upepsilon_n$ is drawn independently from a zero-mean Gaussian distribution with variance $\sigma_\upepsilon^2$. This variance $\sigma_\upepsilon^2$ is a hyperparameter of the experiment, i.e., we will vary it to study its effect on the performance of the proposed approaches. We also center $\vec{\boldsymbol{\upepsilon}}$ by substracting its sample mean value from each entry of the vector. Then, an observation value $\vec{\mathbf{y}}$ is generated from the model~\eqref{eq:GLM-centered}, i.e., 
\begin{equation}
	\vec{\mathbf{y}} =  \vec{\mathbf{x}}_{:,1} \, (\theta_{1} - \theta_{2}) +  \vec{\boldsymbol{\upepsilon}},
\end{equation}
where $\vec{\mathbf{x}}_{:,1} \in {\mathbb R}^{N}$ has half of its elements equal to $-1/2$ and the other half equal to $1/2$. For simplicity, we set $\theta_1 - \theta_2 = 1$. Consequently, the ideal regressor should give out $w_\text{ideal} = 1/(\theta_1 - \theta_2) = 1$. Figure~\ref{fig:error-criteria} represents the estimation error obtained by using criteria~\eqref{eq:MSEcriterion} and~\eqref{eq:L1criterion} as a function of $w$ for different values of $\sigma_\upepsilon^2$. Each curve is the average of $100$ independent experiments. It is seen that both criteria are equivalent for low-noise levels, but the $L_1$ norm based approach is more robust when the noise increases. This finding supports our preference for the use of a regression method based on the $L_1$ norm (such as SVR) in the present research.

\end{document}